%% file: main.tex
\documentclass{article}

\usepackage{microtype}
\usepackage{graphicx}
\usepackage{subfigure}
\usepackage{booktabs} 

\usepackage{hyperref}

\usepackage[preprint]{icml2020}

\usepackage[utf8]{inputenc} 
\usepackage[T1]{fontenc}    
\usepackage{hyperref}       
\usepackage{url}            
\usepackage{booktabs}       
\usepackage{amsfonts}       
\usepackage{nicefrac}       
\usepackage{microtype}      

\usepackage[utf8]{inputenc} 
\usepackage[T1]{fontenc}    
\usepackage{hyperref}       
\usepackage{url}            
\usepackage{booktabs}       
\usepackage{amsfonts}       
\usepackage{nicefrac}       
\usepackage{microtype}      

\usepackage{graphicx}
\usepackage{comment}
\usepackage{enumitem}
\usepackage{wrapfig}

\usepackage{tikz-qtree}
\usepackage{tikz}
\usetikzlibrary{bayesnet}
\usetikzlibrary{arrows}

\usepackage{amsmath,amsfonts,amssymb,amsthm}
\usepackage{mathtools}
\usepackage{commath}
\newtheoremstyle{break}
  {}
  {}
  {\itshape}
  {}
  {\bfseries}
  {.}
  {\newline}
  {}

\theoremstyle{break}
\newtheorem{definition}{Definition}

\newcommand{\Y}{\mathcal{Y}}
\newcommand{\E}{\mathbb{E}}

\newcommand{\cace}{\text{CaCE }}

\icmltitlerunning{Explaining Classifiers with Causal Concept Effect (CaCE)}

\begin{document}

\twocolumn[
\icmltitle{Explaining Classifiers with Causal Concept Effect (CaCE)}



\icmlsetsymbol{equal}{*}
\icmlsetsymbol{note}{*}

\begin{icmlauthorlist}
\icmlauthor{Yash Goyal}{geo,note}
\icmlauthor{Amir Feder}{tech}
\icmlauthor{Uri Shalit}{tech}
\icmlauthor{Been Kim}{goo}
\end{icmlauthorlist}

\icmlaffiliation{geo}{Georgia Tech}
\icmlaffiliation{tech}{Technion}
\icmlaffiliation{goo}{Google Brain}

\icmlcorrespondingauthor{Yash Goyal}{ygoyal@gatech.edu}

\icmlkeywords{Interpretability, Causal Inference, Deep Learning}

\vskip 0.3in
]



\printAffiliationsAndNotice{}  

\begin{abstract}
\input{abstract}
\end{abstract}

\input{chapters/introduction}
\input{chapters/related}
\input{chapters/cace}
\input{chapters/method}

\input{chapters/results}
\input{chapters/conclusion}


\bibliographystyle{icml2020}
\bibliography{cace}

\end{document}

%% file: abstract.tex
How can we understand classification decisions made by deep neural networks? Many existing explainability methods rely solely on correlations and fail to account for confounding, which may result in potentially misleading explanations. To overcome this problem, we define the Causal Concept Effect (CaCE) as the causal effect of (the presence or absence of) a human-interpretable concept on a deep neural net’s predictions. We show that the CaCE measure can avoid errors stemming from confounding. Estimating CaCE is difficult in situations where we cannot easily simulate the do-operator. To mitigate this problem, we use a generative model, specifically a Variational AutoEncoder (VAE), to measure VAE-CaCE. In an extensive experimental analysis, we show that the VAE-CaCE is able to estimate the true concept causal effect, compared to baselines for a number of datasets including high dimensional images. 

%% file: chapters/introduction.tex
\section{Introduction}
\label{sec:intro}

The rise of machine learning use in many practical applications has brought up a new challenge: how to interpret and understand the reasons behind a model's prediction. Particularly in high-risk domains such as medicine, security, etc., it has been widely recognized that understanding the model's reasoning for a prediction is one of the crucial components for wide and safe adoption of the technology. 

The machine learning community has been responding to this demand. Many approaches have been proposed to tackle the challenge: for example by developing a model with interpretable components built-in~\cite{kim2014bayesian, UstunRu14} or by building post-training interpretability methods~\cite{LIME, smilkov2017smoothgrad, fong2017interpretable, dabkowski2017real,chang2018explaining, kim2018interpretability}. While these methods may be useful in showing features or concepts that are correlated with a model's prediction, their explanations might be confounded by correlations present in the data which are not causally relevant to the model, as we describe now.

Say we wish to explain what drives classification decisions for an entire class by a deep neural network, e.g. ``what drives the decision to classify an image as \textsc{bicycle}?''. Now consider the following case: within the training dataset, there is $0.8$ correlation between the presence of cars and the presence of bicycles. However, the dataset is diverse enough and the classifier is powerful enough such that it does \emph{not} rely on the presence of cars in order to classify bicycles: If we were to take images containing bicycles and edit out the cars, we will find that the classifier's output for the label \textsc{bicycle} is virtually unchanged. Even so, as we show below, the strong correlation between cars and bicycles can lead many interpretability methods to wrongfully give the concept \textsc{car} as an explanation for classifying bicycles. In this work, we attempt to tease out the causal aspect: does the presence of a concept like \textsc{car} actually change the classifier's output. The case of ``editing out the cars'' is an example of what is known as the \emph{do}-operator \cite{pearl2009causality}: it formalizes the act of intervening in the world, an act which lies in the heart of defining and understanding causal effects.

In this paper we propose explaining classifiers with the Causal Concept Effect (CaCE) for high-level concepts whose presence or absence (everything else being equal) affect
the model's prediction, as opposed to merely being correlated with the model's prediction. 
CaCE is a global explanation method, where the goal is to explain a model's prediction for an entire class, rather than individual data points (i.e., local explanation methods). As global methods aim to summarize all data points, they are much more vulnerable to confounding of concepts. By concept, we mean a higher level unit than low level, individual input features such as pixels. The example of the cars and bicycles above illustrates the issue. Concepts are often highly correlated with each other in datasets, and we want our explanations to zero-in on the concepts whose presence or absence in isolation causally affects the model's output.

One of the challenges in estimating CaCE for high dimensional data such as images is that there is often no easy way to directly perform the intervention of adding or removing a concept. In the example above regarding \textsc{cars} and \textsc{bicycles}, we might consider ways to cut and/or paste the object in the image. For a more challenging scenario, consider the case where we want to know the causal effect of the concept \textsc{male} on a classifier for \textsc{doctor}. Here, the ideal intervention would be to replace all male doctors in an image with female doctors who are dressed and positioned in the same way as the male doctor, and examine the change in the classifiers output. This is a difficult task, but we  claim that it is worthwhile to try and approximate it using generative models. We show that this approach can successfully capture at least part of the actual causal effect of the concept on the classifier's decisions.

Specifically, we propose using conditional VAEs \cite{cvae, lorberbom2019direct} trained on the training data of the classifier of interest to approximate the true generation process conditioned on concepts. 
Interventions in the generation process then translate to using different values of concepts as inputs to the generative network of the conditional VAE.
We show our approach can approximate CaCE well for a number of datasets.

Our main contributions are the following:
\begin{itemize}[leftmargin=*,topsep=0pt,itemsep=-1ex,partopsep=1ex,parsep=1ex]
\item We propose a general framework to quantitatively measure the causal effect of concept explanations on a deep model’s prediction.
\item We propose an approach which uses conditional generative models to generate counterfactuals and approximate the causal effect of concept explanations.
\item We demonstrate the effectiveness of our approach in estimating the true causal effect of concept explanations for a number of datasets including high dimensional images.
\end{itemize}

%% file: chapters/related.tex
\section{Related Work}
\label{sec:related}

The relationship between causality and explainability has a long history, see \cite{woodward2005making} for a discussion from a philosophy of science point of view. \citet{halpern2005causes} give a formal causal theory of what constitutes an explanation, in terms of what is known as ``actual causality''. In this paper we use a more focussed notion of explanation, i.e., we use the causal effect of a concept on the output of a given trained model as a form of explanation in and of itself.

Recent interpretability methods typically fall into two categories: \emph{global} or \emph{local}~\cite{DoshiKim2017Interpretability}. Global methods explain how a model classifies the predictions for an entire class. Local methods explain how a model classifies a single test instance and answer questions such as ``which part of the test input is most responsible for the classification output?''. While local explanations are important for investigating predictions for individual data points, global explanations are more informative in evaluating the overall robustness of the model and in making deployment decisions.

Our work aims to improve the limitation of global explanation methods: the problem of confounded concepts, and proposes a solution towards correcting it. Specifically, we focus on a recent work TCAV~\cite{kim2018interpretability} that generates high-level concept-based global explanations. It may work well in a variety of applications, however, since various concepts (such as cars and roads) are often highly correlated with each other in the data, TCAV suffers from this confounding of concepts, often providing potentially misleading explanations.
Ideally, we want our explanations to identify the concepts which \textit{causally} affect the model's output.
In this work, we take a step in this direction by formally defining the causal effect of a concept explanation on a model's output and proposing an approach to estimate it.

We note that this problem does not exist as such for most local interpretation methods: because for a given image, the pixels deterministically cause the output of a model, there is no notion of probability or confounding. However, confounding might affect local models where pixels are perturbed based on data-dependent models (e.g. \cite{dabkowski2017real,fong2017interpretable,chang2018explaining}). We leave these cases for future work.

Many interpretability methods developed to have causal-flavor are for \textit{local} explanations, such as removing and adding pixels to generate counterfactual explanations for images~\cite{goyal2019counterfactual, chang2018explaining} or for texts~\cite{hendricks2018generating}. In particular, \cite{chang2018explaining} used the language of counterfactuals to generate \textit{local} explanations. In addition to local and global differences, our work and these prior works face different sets of challenges:
performing the \emph{do}-operation with pixels merely involves changing specific pixels. However, the space of possible operations (combinations of pixels) is huge, as there are millions of pixels, each attaining one of hundreds of values. On other hand, realizing~\emph{do}-operation on concepts is not trivial as it requires some form of data generation process; it is no longer just about changing specific pixels.
However, the space of possible operations is much smaller than that on pixels. Our goal is to generate global concept-based explanations that can succinctly explain if the presence or absence of concepts cause the model's prediction or not.

%% file: chapters/cace.tex
\section{CaCE: Causal Concept Effect\label{sec:cace}}

\begin{figure}[htbp]
 \centering
 \includegraphics[width=.9\linewidth]{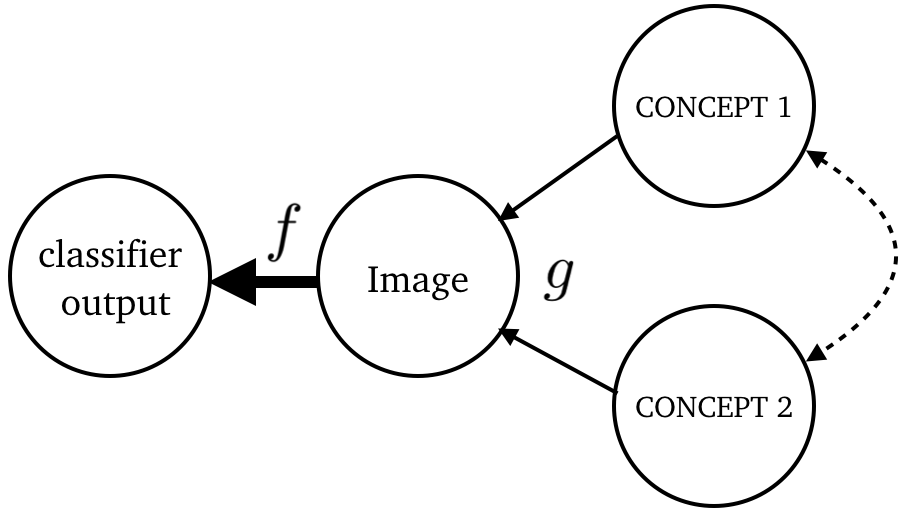}    
    \caption{Causal graph relating high-level concepts (such as objects, background, lighting conditions, etc.), images and classifier $f$ output. The dashed edge indicates possible confounding of the two concepts, by other concepts (not shown in the graph). The thick arrow from Image to the output of $f$ indicates that this relation is mechanistic and we have direct access to it through our knowledge of $f$. This is different from the edges connecting the concepts to the Image, which correspond to the natural generation process of images. In Section \ref{sec:meth}, we propose using a conditional-VAE conditioned on concepts to approximate this relation.}
    \vskip -10pt
    \label{fig:cg1}    
\end{figure}

Denote by $I$ an image, and let $f: I \rightarrow \mathcal{Y}$ be a trained classification model whose output we wish to explain\footnote{For a binary classifier, we typically have $\Y = [0,1]$.}. Let $C_0, \ldots C_k$ be concepts which are potential causes of an image: these may be objects such as ``cars'' or ``bicycles'', but also more abstract concepts such as  ``night time'' or ``brightness above some threshold''.

Let's consider the process that gives rise to the pixels of a typical natural image. We build upon the idea noted in a previous work \cite{ScholkopfJPSZMJ2012} that the causal generation process for a digit image proceeds as the following: a person intends to write the digit $7$, say, and this intention causes a motor pattern producing an image of the digit $7$ -- in that sense the class label $7$ causes the digit image. Generalizing this idea to a natural image, there are many objects in the world, different backgrounds, lighting and angle decisions, as well as properties of the camera and signal processing, all these concepts lead to an ordered set of pixels which is an image.
These concepts are all considered as \emph{causes} for the image, as shown in a simplified example in Figure \ref{fig:cg1}. We say they cause the image since there is a mechanism at work that, given all these concepts, creates a distribution of images with the relevant concepts. Importantly, concepts are far from independent, as some objects typically occur together within a given set of images, as in the car and bicycle example above. 

Note that for a fixed classifier $f$, the only dependence of its output on the concepts is through the image itself. Moreover, the mechanism leading from image to classifier output is in principle known to us, because we have access to the model $f$. On the other hand, our understanding of the generative process $g$ leading from concepts to the image can vary. For example, if we take $g$ to be the natural process giving rise to images of streets with cars and bicycles, it might be very complex. If we take $g$ to be a controlled process whereby we paste a fixed image of a car onto an existing image, then $g$ is relatively simpler. 
To make this distinction between $f$ and $g$, in Figure \ref{fig:cg1}, we represent the potentially complex image generative process $g$ by thin arrows from the Concept nodes to the Image node, and the fixed classifier mechanism by a thick arrow from Image to Classifier Output.

Let $C_0, \ldots, C_k$ be a set of concepts representing the causes of images and $g$ be the generative process giving rise to images. For simplicity, we assume the concepts to be binary corresponding to the presence or absence of the concept.
We define the following structural causal model (SCM, \citet{pearl2009causality}) for the image $I$:
\begin{align}\label{eq:scm1}
    & (C_0, C_1, \ldots, C_k) =  h(\epsilon_C) \nonumber \\
    & I  = g(C_0, C_1, \ldots, C_k,\epsilon_I),
\end{align}
where, as is standard in SCMs, $\epsilon_C$ and $\epsilon_I$ are independent ``noise'' variables. The function $h$ is the  generation process of the concept variables from the random variable $\epsilon_C$ and is not the focus of this work. 
The interventional SCM setting $C_0$ to $a\in\{0,1\}$ is then 
\begin{align}\label{eq:scm2}
    & (C_0, C_1, \ldots, C_k) =  h(\epsilon_C) \nonumber \\
    & C_0 = a \nonumber \\
    & I  = g(C_0, C_1, \ldots, C_k,\epsilon_I).
\end{align}
The SCMs in Eqs \eqref{eq:scm1} and \eqref{eq:scm2} make an important assumption that it is possible to intervene atomically on one concept while leaving all others the same. This might fail for example if some of the concepts are mutually exclusive.
We denote expectations under the interventional distribution by the standard do-operator notation $\E_g\left[\cdot | do(C_0=a)\right]$, where the subscript $g$ indicates that this expectation also depends on the choice of the generative process $g$.

\begin{definition}[Causal Concept Effect, CaCE] 
The causal effect of a binary concept $C_0$ on the output of the classifier $f$ under the generative process $g$ is:
\begin{align*}
    &\text{CaCE}(C_0,f) = \\ &\E_g\left[f(I)|do(C_0=1)\right] 
    - \E_g\left[f(I)|do(C_0=0)\right] .
\end{align*}
\end{definition}
In causal inference literature this is simply known as the average treatment effect (ATE) or average causal effect (ACE) of the concept $C_0$ on the output $f(I)$. We use the term CaCE to focus the discussion on explaining an image classification model's outputs in terms of concepts.

\subsection{CaCE for $N$-way categorical concepts}
\label{sec:mnist_cace}

We can further generalize our definition for CaCE for $N$-way categorical concept variable.
To account for more than 2 possible values of the concept of interest $C_0$, we define \cace for a pair of concept values $C_0=a$ and $C_0=b$ as the following: 

\begin{align*}
    &CaCE(C_0,f, a, b) = \\ &\E_g\left[f(I)|do(C_0=a)\right] 
    - \E_g\left[f(I)|do(C_0=b)\right] 
\end{align*}

If $C_0$ is binary, $a=1$ and $b=0$ are the most obvious choices resulting in \textit{Definition 1}.
If $C_0$ is $N$-way categorical, we treat $b$ as the base value of the concept $C_0$ and marginalize $a$ over all other possible values of $C_0$. Hence\footnote{subscript $g$ has been omitted for simplicity.}, 

\begin{align*}
    &CaCE(C_0,f, b) = \\ \frac{1}{N} \sum_{n \in \set{N} \setminus \set{b}} &\E\left[f(I)|do(C_0=n)\right] 
    - \E\left[f(I)|do(C_0=b)\right] 
\end{align*}

where $\set{N} = \{1, 2, ..., N\}$ refers to the set of all possible values of $C_0$.
Now, for a given image $I_i$, we can calculate \cace using the above definition by fixing $b$ to be the concept label of $I_i$.

\subsection{Why do we need CaCE? }

The first question that comes to mind is -- ``Why do we need CaCE?''.
Let us address this question using a simple example where we can directly control the image generation process.
Lets consider an image classification dataset where an image contains exactly one bar, as shown in Figure~\ref{fig:dec_cace}. 
The orientation of the bar can either be horizontal or vertical, which defines the class label of the image -- $0$ for horizontal and $1$ for vertical.
The bar can also take two different colors -- red or green. In this case, we consider the color as a binary concept ($0$ for red and $1$ for green) for which we measure our metric ``CaCE''. We train a network to classify the images according to the class labels, in two different scenarios. In both scenarios, we can calculate the CaCE exactly, by intervening directly on the images and changing the color of the bars while keeping all other things (location of the bar, its width, etc.) fixed. For each image in the test set, we compute the difference in the classifier's outputs for the original test image and for the counterfactual image with the flipped color. CaCE, by definition, is the average of these differences (each with appropriate $+$ or $-$ sign).

We first consider an \emph{unconfounded} scenario when 
each concept (red or green) is equally balanced with the labels (horizontal and vertical). In other words,  
each class contains equal number of red and green colored bars. In this case, CaCE for the color concept turns out to be zero, as expected. The network learns to ignore the irrelevant concept (color) while making its predictions.  TCAV \citep{kim2018interpretability}, a correlation-based interpretability method, also scores the importance of color concept as zero, i.e., the color concept does not explain the model's decisions, as we would expect.

In a more interesting scenario, we consider a biased dataset. In this case, 90\% of the horizontal bars are red in color (and 10\% are green) and only 10\% of the vertical bars are red in color (and 90\% are green). Unlike the previous scenario, color and class are strongly confounded in this dataset. However, we find that with enough training data, the network still learns to ignore the color; it makes the same prediction for a pair of images which differ from each other only in terms of the color of the bar. Hence, the CaCE of the color concept is still zero. However, we find that TCAV score for the color concept in this binary classification is 1.0 probably because TCAV is fooled by the strong correlation between the class label and the concept, even though the network itself learned to ignore the color.
Hence, there is a need for a mechanism to measure the causal effect of such explanations.

%% file: chapters/method.tex
\section{Measuring CaCE}
\label{sec:meth}

Estimating CaCE in a real-world setting is an exceedingly challenging task: we do not have a good understanding of the complete causal graph underlying the probability distribution of natural images. This lack of understanding implies, for example, that we cannot confidently use standard causal inference methods such as backdoor adjustment. The reason is that in most cases we cannot hope to account for all the confounding factors between a concept and an image. Note also that the image pixels are causally downstream from the concepts, so we cannot simply condition on them.

However, we believe that this fact should not lead us to completely give up on trying to limit confounding in explanations. We focus on approaches that allow us to perform or approximate the $do$-operator directly on images. Thus, our major challenge is instantiating the intervention $do(C_0=a)$ for an image generative process $g$. For example, consider the quantity $\E_g\left[f(I)|do(\textsc{bicycle}=1\right)]$: how can we intervene on an image so that it has a bicycle? 

We present the following two approaches for measuring CaCE: 
(1) Using controlled environments where we can directly intervene on parts of the image generation process
and calculate the exact value of CaCE for this intervention, and (2) Using image generative models (\citep{lorberbom2019direct}, \citep{kocaoglu2017causalgan}) -- while we do not have a full causal graph underlying image generation, we can try and leverage the great strides made in recent years in generative modeling. This approach is not guaranteed to yield the true CaCE value, since the learned generative process is only an approximation of the true generative process tying concepts to images. We therefore conduct a wide set of experiments assessing if and when can the VAE-based approach yield explanations which are less confounded than existing methods. We do this by comparing the VAE-CaCE results to ground-truth CaCE values obtained via interventions in the generation process in controlled settings; by examining and comparing the VAE-CaCE results to other explanation methods; and by proposing and evaluating a set of diagnostic tests whose purpose is to boost our confidence in our results.

\subsection{Ground Truth CaCE (GT-CaCE)}
\label{sec:true_cace}

CaCE can be computed exactly if we can precisely intervene in the generation process of the data.
We call this measure the Ground truth CaCE (\emph{GT-CaCE} in short). 
Under this ideal scenario we can generate the ``true'' counterfactual image for any given image by only changing the concept of interest,
while keeping all the other things in the image fixed.
Then, GT-CaCE can be computed as the mean difference between the predictions of $f$ on these pairs of an image and its counterfactual. 

Since this is a limited scenario and most generation processes we wish to study do not allow such direct interventions, we propose below an approach that learns a generative model approximating the natural image generation process, allowing us to compute an approximation of the CaCE.

\subsection{VAE-CaCE}

We now describe how we use a conditional-VAE to approximate the image generation process and estimate CaCE. The challenge with the VAE approach is that the VAE model is also vulnerable to confounding between the concepts, for example, generating cars when conditioning on bicycles. Generally we cannot guarantee that the learned VAE correctly approximates the true causal generative process of images in the world; however we do believe it captures important parts of it, allowing us to alleviate confounding, while probably not removing it completely. In order to gain confidence in this approach we use extensive experimental evaluation to assess how well does the VAE approach work in scenarios where we do have ground truth. We also propose and conduct a series of diagnostic tests for increasing our confidence in the model's performance. We note that as usual in causal questions, complete confidence is hard to come by.

We choose a conditional generative model because it allows us to generate images for a given concept value $C_0=a$, leaving all other concepts fixed. In addition to conditioning on the concept label, we condition on the class label to be able to generate class-conditional CaCE and to learn a better generative model in comparison to not conditioning on class label. 
Hence, our conditional VAE approximates the distribution $p(I|C_0=a, L=l)$ where $L=l$ denotes the label.

In our experiments, we use the discrete conditional VAE (DC-VAE)\cite{lorberbom2019direct}, specifically their mixture model variant, to allow for discrete latent space. The DC-VAE encoder is comprised of three parts. It has a shared encoder that takes in an image, the class label and the concept label as inputs. Its output is then fed into a discrete and a continuous encoders, resulting in a continuous and a discrete latent spaces.
The decoder takes in samples from both the discrete and continuous latent spaces alongwith the class and the concept labels as inputs, and generates the image reconstruction. For more details, please refer to \cite{lorberbom2019direct}.

We propose two methods of calculating VAE-CaCE: 1) \textit{Dec-CaCE} which only utilizes the generative network (DECoder) of the VAE, and 2)  \textit{EncDec-CaCE} which utilizes both the inference (ENCoder) and the generative networks (DECoder). We describe the two methods in detail below.

\begin{figure}
    \centering
    \includegraphics[width=0.6\linewidth]{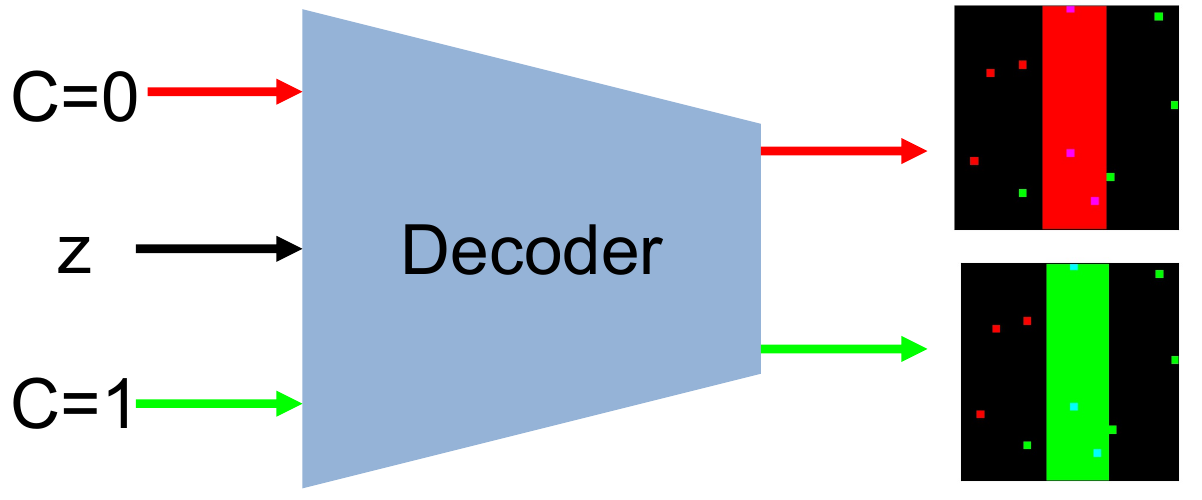}
    \caption{Our Dec-CaCE approach generates pairs of images from the decoder with the same latent sample vector $z$ but with different values of the concept $C$, resulting in pairs of counterfactual images which differ from each other in terms of the concept only.}
    \label{fig:dec_cace}
\end{figure}

\subsubsection{Dec-CaCE}

Analogous to the ideal case of intervening during the generation process (Sec.~\ref{sec:true_cace}), the generative network of the VAE allows us to sample pairs of counterfactual images from $p(I|z,C_0, L)$\footnote{Note that z here refers to the mixed latent space of the VAE for simplicity in notations.} which only differ from each other in the value of the concept $C_0$ (one image for $C_0=0$ and another image for $C_0=1$). This can be achieved by only changing the value $a$ of the concept $C_0$ while keeping the class label $L$ and the sampled latent vector $z$ fixed (see Fig.~\ref{fig:dec_cace}). 
Dec-CaCE is then computed by averaging the difference in the prediction scores $f(I)$ of the two counterfactual images (one with $C_0=1$ and  the other with $C_0=0$) for many random samples of $z$.

\begin{figure}
    \centering
    \includegraphics[width=\linewidth]{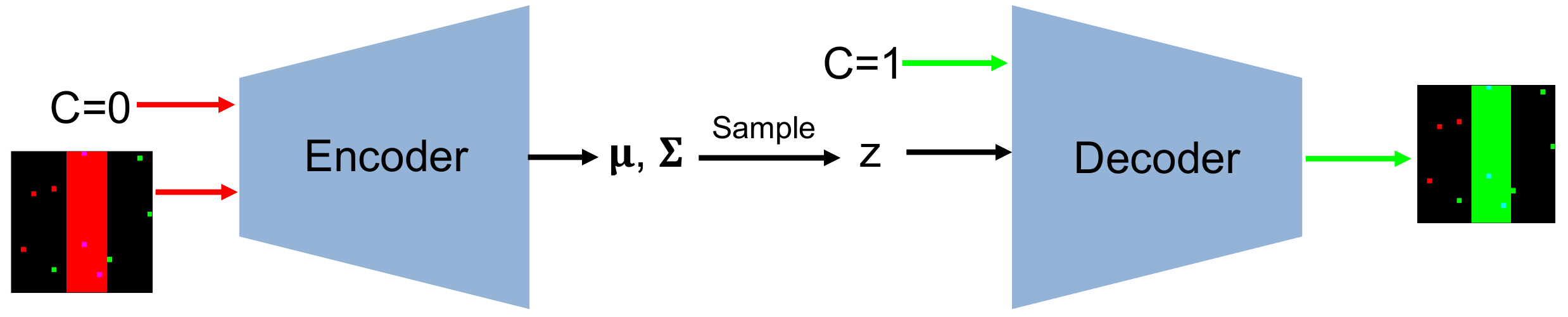}
    \caption{Our EncDec-CaCE approach first infers the latent distribution for a given image and its corresponding concept label using the encoder and then generates a counterfactual image from the decoder using a sample from the inferred latent distribution and the flipped value of the concept $C$.}
    \label{fig:encdec_cace}
\end{figure}

\subsubsection{EncDec-CaCE}

Recall that CaCE is a global explanation method which explains a classifier's predictions for an entire class in terms of a human-interpretable concept. But often times, we might be interested in measuring CaCE for a particular image or a specific set of images, e.g., the set of images for which the classifier makes mistakes. This is not possible using the ``generative-net-only'' approach (Dec-CaCE), but can be achieved by utilizing both the inference and the generative networks of the VAE.

In this approach, for each given image $I_i$ of interest with the class label $l_i$ and the concept label $C_0=c_i$, we can approximately infer the posterior distribution $p(z_i|I_i, L=l_i, C_0 = c_i)$ using the inference network, and then sample a counterfactual image $I_i'$ from $p(I|z_i,L=l_i,C_0=\overline{c_i})$ from the generative network using the flipped value $\overline{c_i}$ of the concept $C_0$ (see Fig.~\ref{fig:encdec_cace}).
EncDec-CaCE is then computed as the difference in the prediction scores of the original image $I_i$ and the generated counterfactual image $I_i'$, i.e., if the concept label $c_i$ for the original image $I_i$ is equal to $1$, $\text{EncDec-CaCE}(I_i) = f(I_i) - f(I_i')$. 
This approach then allows us to estimate CaCE for any set of images by averaging over their individual EncDec-CaCE values.

\subsection{Diagnostic tests on VAE for measuring CaCE}
\label{sec:diganostics}
Our method relies on the assumption that hidden confounding does not significantly impact the concepts and labels we use, and that the VAE successfully disentangles the concept of interest from the other concepts. As is usually the case in causal inference, this assumption is not statistically testable.
We thus propose two simple diagnostic tests for our approach, with the goal of increasing the confidence in our estimates. Note that passing this check does not mean that the estimated CaCE is correct; failing it, however, suggests that the estimated CaCE is probably substantially wrong. We report results from running these tests in Section \ref{sec:diag_results}.
\paragraph{Diagnostic test I: positive effect}
We suggest estimating the CaCE of the label on the classification output of that same label, i.e. we take the concept to be the label itself. Assuming that the classifier $f$ has reasonably good performance classifying the label $l = C_0$, the presence or absence of $l$ should have a strong causal effect on the output. Failure of this diagnostic might mean that the conditional-VAE is weak and does not capture the relation of labels and the image or image embedding.
\paragraph{Diagnostic test II: null effect}
We suggest estimating the CaCE of a concept which we know should have essentially no effect on the output of $f$ (known as ``negative controls'' in the causal inference literature). For example, we can add a random, independent dummy concept with probability $0.5$ to each image in the dataset, and estimate its CaCE.

%% file: chapters/results.tex
\section{Results}
\label{sec:results}

We apply our approach on four datasets -- a synthetically generated BARS dataset, colored-MNIST \cite{kim2019learning}, COCO-Miniplaces \cite{BAM2019} and CelebA \cite{liu2015faceattributes} -- to demonstrate the generalizability of CaCE and our methods for estimating CaCE.
The first three datasets are designed such that we have access to the full data generation process. With such, we can intervene on the concept label of each image to compute the GT-CaCE and evaluate how well our approach works as compared with GT-CaCE.

The definition of \cace outputs a vector with the same dimension as the output of the classifier, i.e., the number of classes.
For binary classification, it is a $2$-dim vector, the two values being equal in magnitude but with opposite signs. In that case, without any loss of information, we report the \cace score for class 0 in our results for BARS, COCO-Miniplaces and CelebA datasets. 
In the case of multiple classes (10-way classification for colored-MNIST dataset), we instead compute the mean absolute difference in the probability distributions, averaged over the classes. 

We compare our results with two baselines:
1) a non-causal baseline with no intervention, which we call \textsc{ConExp}, and 2) TCAV \cite{kim2018interpretability}.
The \textsc{ConExp} simply computes conditional expectation of the prediction scores conditioned on the concept label i.e., same quantity as CaCE but without the do-operators:
$$
    \text{\textsc{ConExp}}(C, f) = \E\left[f(I)|C=1\right] - \E\left[f(I)|C=0\right]
$$
TCAV~\cite{kim2018interpretability} computes a global explanation score for a given (concept, label) pair, purely relying on correlations.
Note that TCAV requires access to internal representations of the image (i.e., embeddings) from the classifier while CaCE and our method VAE-CACE can be applied to any black-box classifier.

\subsection{BARS: CaCE for a synthetic dataset}
\label{sec:bar_results}

We first use a simple dataset of bar images as described in Section~\ref{sec:cace}.
Recall that this dataset is generated such that the orientation of the bar (e.g., vertical v.s. horizontal) indicates the class label, and the color of the bar represents the concept. We create a number of biased datasets by varying how often each concept (color) appears in each class (each row in Table~\ref{tab:cace_toy}) to create a spectrum of values of GT-CaCEs.
For each dataset, we learn a simple binary classifier consisting of 3 fully-connected layers. 

\begin{table}[h!]
  \caption{CaCE results for BARS dataset.}
  \label{tab:cace_toy}
  \centering
            \resizebox{1.\columnwidth}{!}{
  \begin{tabular}{|p{1.9cm}|p{1.8cm}|p{1cm}|p{1cm}|p{1cm}|p{1cm}|p{1cm}|}
    \toprule
    \% of red in class 0 (horz)     & \% of red in class 1 (vert)     & GT-CaCE     & Dec-CaCE   & EncDec-CaCE  & ConExp & TCAV \\
    \midrule
    60  & 40    &   0.00  &   0.02    & 0.00  & 0.21    & 0.76 \\
    99  & 01    &   0.58  &   0.59    & 0.30  & 1.00    & 1.00   \\
    98  & 02    &   0.47  &   0.49    & 0.26  & 0.97    & 0.96  \\
    99  & 50    &   0.39  &   0.44    & 0.04  & 0.69    & 0.96  \\
    
    \bottomrule
  \end{tabular}
  }
\end{table}

In a case when the correlation between color and class is low (first row in Table~\ref{tab:cace_toy}), GT-CaCE is zero, meaning that the classifier ignores the color. While our all methods measure close to zero, ConExp and TCAV incorrectly reports higher impact of the color. In contrast, when the correlation between color and class is high (second row in Table~\ref{tab:cace_toy}), GT-CaCE is $0.58$, meaning that the classifier does rely on colors to some extent (note that the highest CaCE is 1.0), Dec-CacE estimates this correctly, while ConExp and TCAV again only report pure correlations. Among our methods, we find that Dec-CaCE performs better than EncDec-CaCE.

\subsection{Colored-MNIST: CaCE for $N$-way categorical concept}
\label{sec:mnist_results}

The colored-MNIST dataset~\cite{kim2019learning} is a variant of the MNIST dataset \cite{deng2012mnist} where the foreground digit is colored. The digit defines the class of the image, and the color defines the concept. Following \cite{kim2019learning}, a color bias is introduced in the dataset in the following way: each digit class is assigned a mean color value. For each image, a color value is sampled from a normal distribution with the digit class's color value as mean and a fixed covariance ($\sigma$, same for all classes). The value of covariance, $\sigma$, can be used to control the degree of the bias in the dataset (smaller $\sigma$ means highly biased). We created a number of biased datasets by varying $\sigma$ from 0.02 to 0.05 with an interval of 0.005. For each dataset, we train a ResNet-100 classifier for digit classification.
Unlike the BARS dataset, the color concept is non-binary; it is defined as the closest out of 13 possible colors. Therefore, CaCE measures in this case are measured as described in Section~\ref{sec:mnist_cace}.

\begin{table}[h!]
  \caption{CaCE results for Colored-MNIST dataset.}
  \label{tab:cace_mnist}
  \centering
            \resizebox{1.\columnwidth}{!}{
  \begin{tabular}{|p{1.0cm}|p{1.5cm}|p{1.6cm}|p{1.5cm}|p{1.5cm}|p{1.cm}|}
    \toprule
    $\sigma$
    & Avg-GT-CaCE & Dec-CaCE & EncDec-CaCE  & ConExp & TCAV \\
    \midrule
    0.02 & 0.094 & 0.097 & 0.099 &0.154&  0.16\\
    0.025 & 0.089 & 0.092 & 0.093 &0.147&  0.155\\
    0.03 & 0.076 & 0.079 & 0.08  &0.135&  0.152\\
    0.035 & 0.068 & 0.07 & 0.071  &0.133&  0.152\\
    0.04 & 0.061 & 0.063 & 0.065  &0.121&0.139\\
    0.045 & 0.062 & 0.065 & 0.066  &0.131&  0.13\\
    0.05 & 0.058 & 0.061 & 0.062  &0.118&  0.117\\
    \bottomrule
  \end{tabular}
  }
\end{table}

\begin{figure}
    \centering
    \includegraphics[width=0.8\linewidth]{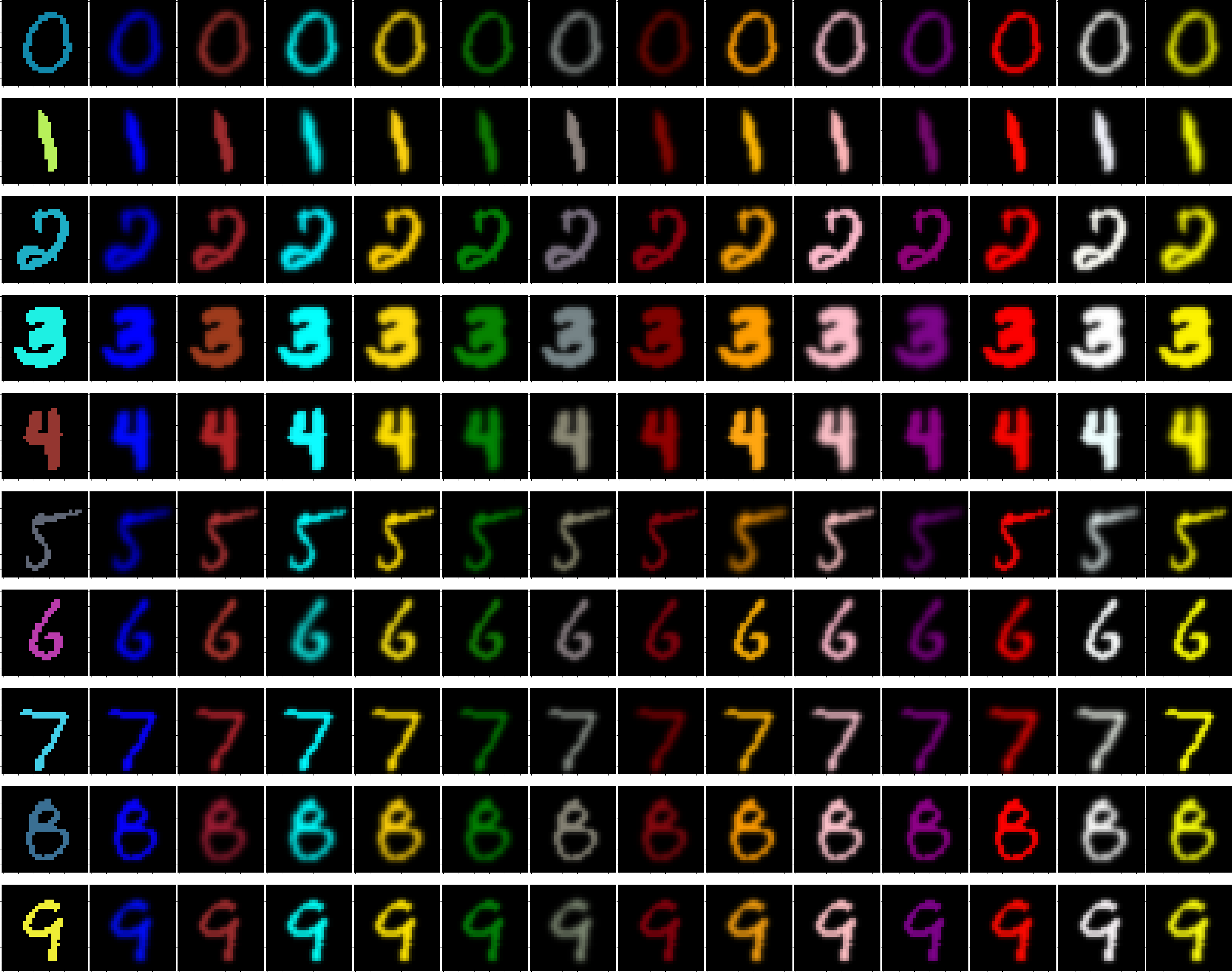}
    \caption{Colored-MNIST images from the test set (leftmost column) alongside their counterfactuals generated from \textit{DC-VAE}. Each of the 2-14 columns correspond to one possible value for the color concept. Each row corresponds to one image from each digit class.}
    \label{fig:mnist}
\end{figure}

As shown in Table~\ref{tab:cace_mnist},
we can see that our methods -- Dec-CaCE and EncDec-CaCE are able to estimate CaCE well while baselines method tend to overestimate the importance of the concept.
Examples of the counterfactual images generated from our DC-VAE are shown in Fig.~\ref{fig:mnist}. 

\paragraph{CaCE estimate as complexity of classifiers vary.}\label{sec:generalization_mnist}

How does CaCE estimate vary as the complexity of the classifier change?
We repeat the experiments from Table~\ref{tab:cace_mnist} (which correspond to ResNet-100) for ResNet-1 and a \textit{simple-CNN} model consisting of 2 convolutional layers and a fc layer.
We find that a more complex classifier (ResNet-100) tends to be more affected by the correlation between class and color concept and results in higher CaCE values as compared to relatively simpler classifiers (such as simple-CNN).

\paragraph{CaCE estimate as complexity of generative models vary.}
We ask similar question now with generative models.
We again repeat the experiments from Table~\ref{tab:cace_mnist} (which correspond to a DC-VAE encoder architecture consisting of convolutional layers) for a simpler DC-VAE encoder architecture consisting of only fc layers. We find that while both DC-VAEs can estimate CaCE pretty well, the estimates from ``convolutional'' DC-VAE are $8$-$49\%$ closer to GT-CaCE as compared to ``fc'' DC-VAE.

\subsection{COCO-Miniplaces: CaCE for high-dimensional images}
\label{sec:dogscenes_results}
This dataset, introduced in \cite{BAM2019}, combines images from Miniplaces \cite{Zhou17} and COCO \cite{coco} datasets.
Each image in this dataset is generated by pasting an object segmentation crop (e.g., a dog) from a COCO image on to a scene image. 
The Miniplaces scene category defines the class label for the modified image, while the presence or absence of the object crop is a binary concept. 

Similar to the BARS dataset, this dataset allows us to control the value of the binary concept 
during the generation process.
We vary how often the object is pasted on to the images for a class 
to create a range of GT-CaCE values.

We create this dataset to have a simple setup, potentially to control for unrelated randomness and focus on investigating the causal effect of the presence and absence of the objects. We use a single object and keep the size of the crop and its pasted location in the scene image to be fixed. 

To encourage the classifier to use the presence or absence of the object as a signal for its prediction, we manually choose the two most confusing classes from the dataset -- `bathroom' and `shower'. Note that if two classes are not confusing, we observe that classifiers lean to ignore the object, resulting GT-CaCE to be zero.
For each setting of the dataset, we finetune a ResNet-50 model, pretrained for 365-way classification on Places dataset \cite{Zhou17}, for binary classification between these 2 classes.
To train VAEs, we augment the Miniplaces dataset with images from the Places dataset~\cite{Zhou17} to increase the size of the training dataset (note that Miniplaces is a subset of Places dataset).

\begin{table}[h!]
  \caption{CaCE results for COCO-Miniplaces dataset.}
  \label{tab:natural_cace}
  \centering
              \resizebox{1.\columnwidth}{!}{
  \begin{tabular}{|p{1.7cm}|p{1.5cm}|p{1cm}|p{1cm}|p{1cm}|p{1cm}|p{1cm}|}
  \toprule
    \% of obj in `bathroom'     & \% of obj in `shower'     & GT-CaCE     & Dec-CaCE  & EncDec-CaCE & ConExp & TCAV \\
    \midrule
    60  & 40    &   0.13    &   0.154    & 0.078   & 0.23 & 0.723    \\
    99  & 01    &   0.694    &   0.651    & 0.345   & 0.841 & 1.000     \\
    95  & 05    &   0.604    &   0.543    & 0.262   & 0.791 & 0.988    \\
    99  & 50    &   0.328    &   0.31    & 0.291   & 0.49 & 0.944   \\
    \bottomrule
  \end{tabular}
  }
\end{table}

Table~\ref{tab:natural_cace} shows CaCE scores for class 0 (`bathroom').
For the first case where 60\% of the `bathroom' and 40\% of the `shower' images contain the object, the GT-CaCE and estimates from all our methods are small, while $ConExp$ and TCAV incorrectly assign a large importance to the concept, consistent with results in Sec.~\ref{sec:bar_results}.

In cases of high correlation (rows 2-4), we observe that our method Dec-CaCE estimates CaCE values close to the GT-CaCE values in all cases. On the other hand, the baseline $\textsc{ConExp}$ and TCAV tend to overestimate the importance of the concept, as we would expect when the concept is strongly correlated with the label. 
As evident from these empirical results, we believe our CaCE estimates can help provide a better understanding of the degree to which correlated concepts in high-dimensional images actually impact the classifier's output.

\subsection{CelebA: CaCE for naturally occurring confounding concepts}
\label{sec:celeba_results}

Lastly, we showcase our estimates in a real dataset, where confoundings may naturally occur.
CelebA \cite{liu2015faceattributes} dataset contains celebrities' face images. Each image is annotated with a number of attributes such as gender, smiling, wearing glasses, etc. We consider the binary task of classifying these face images into male or female and consider the attributes of `eyeglasses' and `blonde hair' as binary concepts. 
In order to create biased settings, we subsample from the training set to achieve a desired level of bias. We then train a ResNet-100 classification model for each of these biased datasets.

Since we do not control the image generation process, it is not possible to calculate GT-CaCE. However, for one particular concept, hair color, 
StarGAN~\cite{choi2018stargan} has been shown to generate realistic counterfactual images for the blonde hair attribute. We report approximated GT-CaCE using StarGAN by generating realistic counterfactual images for the blonde hair and calculating EncDec-CaCE for all test images (third column in Table~\ref{tab:cace_celeba}).

\begin{table}[h!]
\caption{CaCE results for CelebA dataset for `blonde hair' concept.}
\label{tab:cace_celeba}
  \centering
\resizebox{1.\columnwidth}{!}{
  \begin{tabular}{|p{1.cm}|p{1.cm}|p{1.2cm}|p{1.cm}|p{1.cm}|p{1.cm}|p{1cm}|p{1.0cm}|p{1.0cm}|}
    \toprule
    \% of blonde women & \% of blonde men & StarGAN & Dec-CaCE & EncDec-CaCE & ConExp & TCAV \\
    \midrule
    60 & 40 & 0.049 & 0.057 & 0.05 & 0.152 & 0.755 \\
    99 & 1 & 0.523 & 0.585 & 0.513 & 0.752 & 1 \\
    98 & 2 & 0.448 & 0.537 & 0.44 & 0.706 & 1 \\
    95 & 5 & 0.468 & 0.479 & 0.474 & 0.642 & 0.978 \\
    99 & 50 & 0.176 & 0.209 & 0.174 & 0.376 & 0.953 \\
    \bottomrule
  \end{tabular}
  }
\end{table}

\begin{figure}
    \centering
    \includegraphics[width=.8\linewidth]{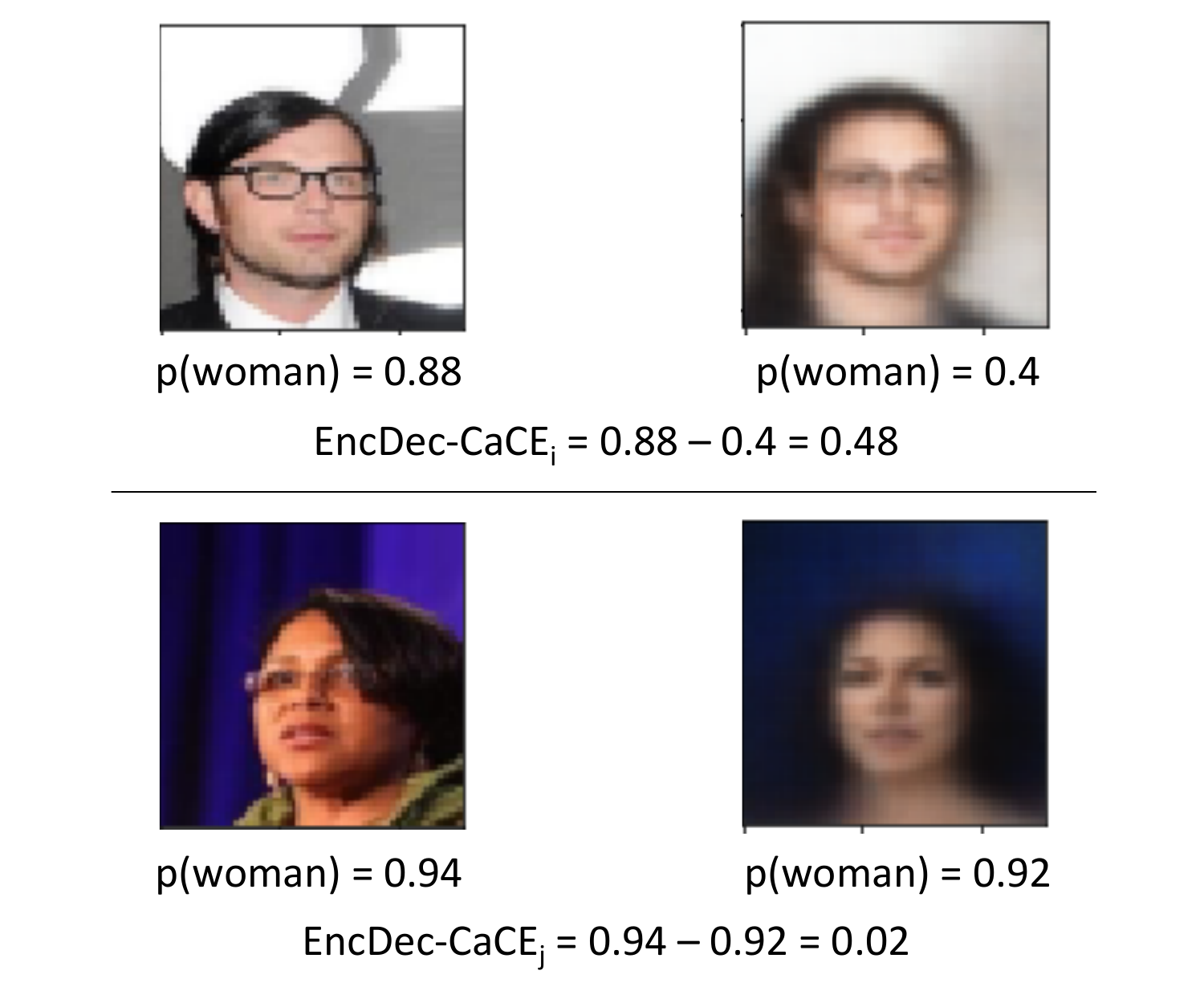}
    \vspace{-8pt}
    \caption{CelebA test images (with eyeglasses) along with their counterfactuals (without eyeglasses) generated from the \textit{DC-VAE}.}
    \label{fig:celeba-vae}
    \vspace{-8pt}
\end{figure}

As shown in Table~\ref{tab:cace_celeba},
we observe consistently similar results for CelebA for the `blonde hair' attribute as we observed for previous three datasets, i.e., we see that our estimates of CaCE are much closer to those obtained by StarGAN, while other baselines do not. 

We also show qualitative examples from our trained VAE for the attribute `eyeglasses' which are used to measure EncDec-CaCE approach in Fig.~\ref{fig:celeba-vae}. For the first example (top row) where the classifier is trained on `95-05' biased dataset (95\% of women images are wearing eyeglasses; 5\% of men images are wearing eyeglasses), the classifier predicts $p(woman)$ = $0.88$ for the test image, perhaps because it has mostly seen eyeglasses in women images (because of the bias in the dataset). It predicts $p(woman)$ = $0.4$ for the generated counterfactual image without eyeglasses (top right). Hence, EncDec-CaCE for this example is $0.48$. The bottom row shows an example from `60-40' biased dataset where the effect of eyeglasses on the classifier's prediction is negligible. Note that EncDec-CaCE numbers reported in the tables are averages of such individual calculations over the entire test set.

\subsection{Diagnostic tests results}
\label{sec:diag_results}

In this section, we report results for the diagnostic tests proposed in Section~\ref{sec:diganostics} for colored-MNIST dataset. For test 1, we set the concept to be each class label and calculate the average CaCE for all classes. For each image in the test set, we generate counterfactual images that differ only in their concept value (in this case, class label), and compare the classifier's outputs.
We observe an average GT-CaCE of $0.152$. Note that we expect this result to be close to the upper limit score ($0.2$).

For test 2, we created a binary dummy concept that appears on each digit image with a probability of $0.5$ across all classes. Since it has no correlation with digit label, the classifier is expected to ignore it, resulting in a CaCE of zero. We observe the CaCE to be $0.003$. Hence, our DC-VAE passes the diagnostics tests. 

%% file: chapters/conclusion.tex
\section{Conclusions}

The goal of interpretability methods is to help humans make decisions about machine learning models, whether the decision is about deployment in high risk domains, or checking if the model is unfair to a subgroup of people. It is critical that the explanations correctly reflect how the model is making predictions, instead of merely reflecting correlations with predictions. We propose a simple metric CaCE, and show it captures more closely what we expect of explanations of models: the causal effect of the absence or presence of a concept on the classifier's output. We then show how we can estimate CaCE, leveraging the recent development of powerful conditional VAEs. We demonstrate that our method can closely match the GT-CaCE for a number of datasets. We hope that CaCE is a starting point towards targeting succinct and causal explanations to unveil the causal processes in classifiers. 